\documentclass[runningheads]{llncs}

\usepackage{eccv}

\usepackage{eccvabbrv}

\usepackage{graphicx}
\usepackage{booktabs}
\usepackage{multirow}
\usepackage[dvipsnames]{xcolor,colortbl}

\usepackage[accsupp]{axessibility}  

\usepackage{hyperref}

\usepackage{orcidlink}

\newcommand{\method}{\textsc{MultiDelete}\xspace}
\newcommand{\propone}{modality decoupling\xspace}
\newcommand{\proptwo}{multimodal knowledge retention\xspace}
\newcommand{\propthree}{unimodal knowledge retention\xspace}

\newcommand{\retrain}{\textsc{Retrain}\xspace}
\newcommand{\ft}{\textsc{FineTune}\xspace}
\newcommand{\neggrad}{\textsc{NegGrad}\xspace}
\newcommand{\dtd}{\textsc{DtD}\xspace}
\newcommand{\erm}{\textsc{Erm-Ktp}\xspace}
\newcommand{\codec}{\textsc{L-codec}\xspace}
\newcommand{\knowul}{\textsc{UL}\xspace}
\newcommand{\dtrain}{$D_{\mathrm{Train}}$\xspace}
\newcommand{\dtest}{$D_{\mathrm{Test}}$\xspace}
\newcommand{\df}{$D_f$\xspace}
\newcommand{\dr}{$D_r$\xspace}
\newcommand{\dfauc}{$D_f | D_r$\xspace}
\definecolor{Gray}{gray}{0.9}

\begin{document}

\title{MultiDelete for Multimodal Machine Unlearning} 


\author{Jiali Cheng \orcidlink{0009-0006-3376-6549} \quad Hadi Amiri}

\authorrunning{J. Cheng and H. Amiri}

\institute{University of Massachusetts Lowell\\
\email{\{jiali\_cheng, hadi\_amiri\}@uml.edu}
}

\maketitle
\begin{abstract}
Machine Unlearning removes specific knowledge about training data samples from an already trained model. 
It has significant practical benefits, such as purging private, inaccurate, or outdated information from trained models without the need for complete re-training. 
Unlearning within a multimodal setting presents unique challenges due to the complex dependencies between different data modalities and the expensive cost of training on large multimodal datasets and architectures. 
This paper presents the first machine unlearning approach for {\em multimodal} data and models, titled \method, which is designed to decouple associations between unimodal data points during unlearning without losing the overall representation strength of the trained model. \method advocates for three key properties for effective multimodal unlearning: 
(a): \propone, which effectively decouples the association between individual unimodal data points marked for deletion, rendering them as unrelated data points,
(b): \proptwo, which retains the multimodal representation post-unlearning, and
(c): \propthree, which retains the unimodal representation post-unlearning.
\method is efficient to train and is not constrained by using a strongly convex loss--a common restriction among existing baselines.
Experiments on two architectures and four datasets, including image-text and graph-text datasets, show that \method 
gains an average improvement of $17.6$ points over best performing baseline in unlearning multimodal samples, 
can maintain the multimodal and unimodal knowledge of the original model post unlearning, and 
can provide better protection to unlearned data against adversarial attacks\footnote{Code and data is available at \url{https://github.com/CLU-UML/MultiDelete}}.

\end{abstract}    
\section{Introduction}
\label{sec:intro}

Multimodal models are used in a variety of real-world applications~\cite{radford2021learning,vilt,albef,pmlr-v162-li22n}, which require extensive training on large datasets. However, the underlying training data can be subject to change due to various reasons such as copyright issues, users revoking consent, and data becoming outdated. Continuing to use a model trained on such data poses significant risks to privacy and questions the model's relevance and accuracy. Machine Unlearning addresses these challenges by removing specific knowledge of training samples from an already trained model, while preserving the model's functionality on its downstream tasks.

Despite recent advances in machine unlearning in {\em unimodal} settings~\cite{Golatkar2020EternalSO,guo-2020-certified-removal,Mehta2022DeepUV,Lin2023ERMKTPKM,jang-etal-2023-knowledge}, unlearning in multimodal settings is largely unexplored. Multimodal unlearning is a challenging task due to the complex relationship and dependency among individual data modalities and the complexity inherent in multimodal data and model architectures. To the best of our knowledge, no existing approach is specifically designed for unlearning in the context of multimodal data, and existing unimodal approaches may not be directly applicable or effective on multimodal data. 
Specifically, existing 
\textit{weight-scrubbing} methods add noise to model weights~\cite{Golatkar2020EternalSO,neel2021descent,Mehta2022DeepUV}, which may fall short in full unlearning of the inter-modality dependencies, which is a key requirement for eliminating any residual data traces. \textit{Certified removal} methods assume convexity of training objectives~\cite{Chien2023EfficientMU,pmlr-v206-cong23a}, which often does not hold in multimodal settings. \textit{Optimization-based} methods focus on unimodal settings or last layer models~\cite{jang-etal-2023-knowledge,Lin2023ERMKTPKM,Liu_2023_ICCV}, which are less effective on multimodal data because of the cross-modality interactions that can occur beyond just the last layer. Finally, \textit{efficient retraining methods} require significant training cost and can lead to overfitting~\cite{bourtoule2021machine,Chen2021GraphU,Dukler_2023_ICCV}.

In this paper, we take the initial step to investigate the multimodal unlearning problem. Our approach to formulating multimodal unlearning is titled \method and centers on developing a model that
satisfies three key properties: 
(a) \propone which reduces the dependencies between modalities for data samples marked for deletion, 
(b) \proptwo which preserves the previously learned multimodal knowledge of the model during unlearning, and 
(c) \propthree which retains the previously learned unimodal knowledge. 
We formally define these properties and design specific loss functions for effective and efficient multimodal unlearning. 
We summarize our contributions as follows:
\begin{itemize}
    \item we conceptualize multimodal unlearning through three pivotal properties: \propone, \proptwo, and \propthree, which collectively enable multimodal unlearning while preserving the essential knowledge previously learned by the model, and
    \item through extensive experiments across image-text and graph-text datasets, and architectures, we show the efficacy of the proposed properties and approach in providing protection to deleted data and maintaining model's robustness against adversarial attacks.
\end{itemize}

Experimental results show the efficacy of \method measured by standard performance metrics across different multimodal tasks, as well as efficiency versus retraining from scratch.
Compared to the best-performing baseline, \method obtains superior performance advantage of $17.6$ in forgetting deleted data, while maintains the previously learned knowledge with $0.3$ performance gain across a wide-range of tasks. In addition, \method can better protect the deleted data against membership inference attacks post-unlearning by $0.13$ absolute points, and is substantially more efficient than retraining the model from scratch. 




\section{\method}
\label{sec:method}
Without loss of generality, we center the presentation of \method on dual data modalities, such as image-text. We note that \method can be applied to more than two modalities or other types of modalities. We demonstrate this broader applicability in experiments using graph-text data.

\subsubsection{Notation}
Consider a vision-language model $f$ trained on a dataset of $N$ image-text pairs, 
$D_{\mathrm{train}} = \{(I_i, T_i)\}^N_{i=1}$. 
We denote $D_f \in D_{\mathrm{train}}$ as the subset of training data that should be unlearned from $f$. Conversely, $D_r = D_{\mathrm{train}} \setminus D_f$ denotes the remaining data after removing $D_f$. We denote $f'$ as the desired unlearned model, which functions as if $D_f$ had never been used in training of $f$.
In addition, we assume that the original model $f$ can be decomposed into a vision feature extractor $f_I$, a language feature extractor $f_T$, and a modality fusion module $f_F$.


\subsubsection{Problem Formulation} Given a vision-language model $f$ trained on $D_{train}$, we aim to unlearn a subset of training data $D_f$ from $f$ and obtain a corresponding unlearned model $f'$.
%
For this purpose, our core objective is to eliminate the influence that any $(I_i, T_i)\in D_f$ has on the parameters of $f$, so that it effectively ``forgets'' the patterns learned from $D_f$. Crucially, we aim to maintain the performance of the unlearned model $f'$ on the test data $D_{test}$ as close as possible to that of the original model $f$. This will ensure that while $f'$ discards specific data knowledge from $D_f$, it retains the overall task effectiveness of $f$ on \dr samples. Ideally, $f'$ should effectively forget \df without compromising its performance on \dtest.
Here, we assume that the model only needs to unlearn the {\em relationship} between image-text pairs $(I_i, T_i) \in D_f$, but not necessarily the individual unimodal elements, $I_i$ and $T_i$ themselves. 
This approach ensures that the model retains its foundational knowledge of individual modalities, which is essential for effective learning of the target task and prevents the unnecessary loss of information. For example, it allows for forgetting of a user's ``like'' on a social media post without requiring to remove either the user or the post from the platform.\looseness-1



\begin{figure*}[t]
    \centering
    \includegraphics[width=0.88\textwidth]{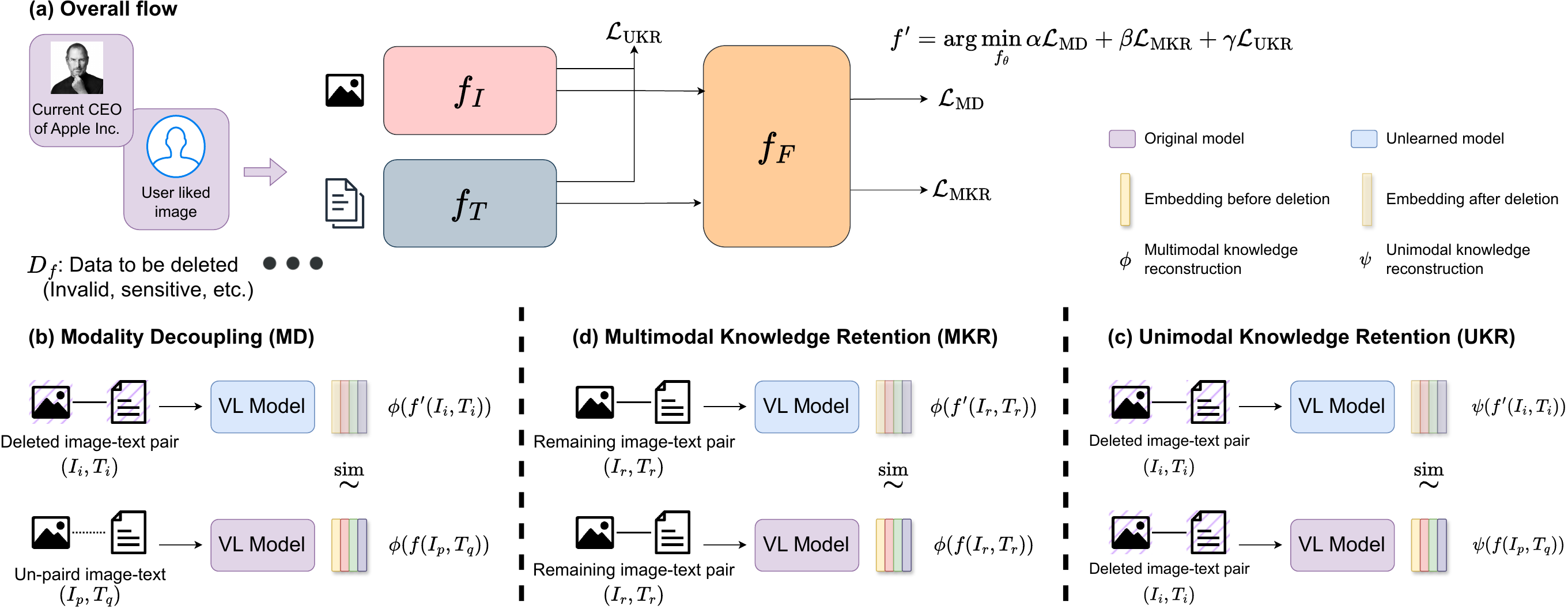}
    \caption{Summary of the proposed approach, \method. 
    \textbf{(a)} given a trained multimodal model (e.g. a vision-language model) and a subset of its training data $D_f$ with image-text {\em relations} marked for unlearning or deletion, \method decouples the inter-modality dependency on $D_f$,
    while maintaining the multimodal and unimodal knowledge on the remaining training data $D_r = D_{\mathrm{train}} \setminus D_f$. 
    \textbf{(b)} \propone: ensures that individual modalities in the deleted data pairs $D_f$ are treated as unrelated by the model,
    \textbf{(c)} \proptwo: preserves the model's overall multimodal knowledge, and
    \textbf{(d)} \propthree: preserves the model's understanding of individual modalities. \vspace{-5mm}
    }
    \label{fig:model}
\end{figure*}

\subsection{Key Properties for Multimodal Unlearning}
\label{sec:properties}

We outline key properties essential for successful multimodal unlearning. These properties and the overall work flow of the model are depicted in \cref{fig:model}.  




\subsubsection{Modality Decoupling} 
This property requires that for the unlearned model $f'$, the relationship between any image-text pair $(I_i, T_i)\in D_f$ should be indistinguishable from the relationship of any non-related image-text pair in the dataset. In other words, $f'$ should be unable to discern a removed pair $(I_i, T_i) \in D_f$ from those that are deemed unassociated in the dataset. 
This decoupling is pivotal because it effectively prevents the reconstruction or extraction of any specific information about the removed pair $(I_i, T_i) \in D_f$. It also helps model's generalizability and robustness, because residual associations from the removed data could potentially skew the model's performance on new data~\cite{Golatkar2020EternalSO}. 
We formally define this notion of unlearning in the multimodal context as follows:


\begin{definition}[\propone]
    Let $(I_i, T_i) \in D_f$ denote an image-text pair marked for deletion from model $f$. The unlearned model $f'$ achieves effective modality decoupling when $(I_i, T_i) \in D_f$ becomes indistinguishable from any unrelated image-text pair $(I_p, T_q), p\neq q$ sourced from $D_r$:
    \begin{equation}
        \mathop{\mathbb{E}}_{(I_i, T_i) \in D_f, (I_p, T_q)_{p \neq q}}\Big[\phi\big(f'(I_i, T_i)\big) - \phi\big(f(I_p, T_q)\big)\Big] = \epsilon,
        \label{eq:propone}
    \end{equation}
where $f(\cdot)$ and $f'(\cdot)$ generate multimodal representations of their inputs, $\phi$ is a readout function (such as the concatenation operator, applied to a set of representations), and $\epsilon$ is an infinitesimal constant. 
\end{definition}


To realize this property, we randomly draw unrelated image text pairs $(I_p, T_q)$ from $D_r$, and minimize the difference in multimodal associations between the image-text pairs $(I_i, T_i)\in D_f$ and the unassociated image-text pairs $(I_p, T_q)$. This is achieved by minimizing the following distance:
\begin{multline}
    \mathcal{L}_{\text{MD}} = \text{Dis}
    \Big(
    \big\{ f'(I_i, T_i) | (I_i, T_i) \in D_f \big\},
    \\ 
    \big\{ f(I_p, T_q) | (I_p, T_p) \in D_r, (I_q, T_q) \in D_r, p \neq q\big\}
    \Big),
\end{multline}
where $\text{Dis}(\cdot)$ can be mean squared error. 
By minimizing this loss, the model is trained to forget or unlearn specific relationships in deleted pairs, making them indistinguishable from unrelated or random data pairs. This is a crucial step in ensuring that the unlearned model does not retain any knowledge about the data it is supposed to forget.

\subsubsection{Multimodal Knowledge Retention} 
This property requires that, the process of unlearning $D_f$ does not adversely affect the learned multimodal knowledge on the remaining dataset $D_r$. In other words, the multimodal knowledge related to image-text pairs in $D_r$, i.e. $f'(I_r, T_r), \forall (I_r, T_r)\in D_r$, should preserve the corresponding original knowledge, $f(I_r, T_r)$, after the unlearning process. This approach ensures that while specific data pairs are being unlearned, the overall multimodal knowledge and capability of the model remain robust.
Formally, we define retention of multimodal knowledge as follows:
\begin{definition}[\proptwo]
        Let $(I_r, T_r) \in D_r$ denote an image-text pair that is ``not'' marked for deletion. The unlearning approach is effective in retaining multimodal knowledge if it minimizes the deviation in the multimodal knowledge between the unlearned model $f'$ and the original model $f$:
    \begin{equation}
        \displaystyle \mathop{\mathbb{E}}_{(I_r, T_r) \in D_r}
        \Big[
        \phi \big(f'(I_r, T_r)\big) - 
        \phi \big(f(I_r, T_r)\big)
        \Big] = \epsilon,
        \label{eq:proptwo}
    \end{equation}
\end{definition} 
\noindent where the readout function $\phi$ is a vector combination operator (such as concatenation). 
We realize this property by minimizing the gap in the multimodal knowledge between $f'$ and $f$ as follows:
\begin{equation}
    \mathcal{L}_{\text{MKR}} = \text{Dis} \Big( f'(I_r, T_r), f(I_r, T_r) \Big), (I_r, T_r) \in D_r.
\end{equation}

\subsubsection{Unimodal Knowledge Retention} 
This property requires that the individual unimodal representations of the data points $(I_i, T_i)\in D_f$ remain intact post unlearning. The rationale is that although the inter-modal relationships are unlearned through modality decoupling, $I_i$ and $T_i$ are still valid standalone image and text data. Therefore, it is important that their unimodal representations are preserved to retain the unimodal knowledge initially learned by $f$, i.e. $f_I(I)$ for images and $f_T(T)$ for texts. 
This property helps maintain the core knowledge of individual modalities, and prevents unnecessary loss of information or the need to relearn basic features from scratch post-unlearning. 
Formally, we define \propthree as follows:
\begin{definition}[\propthree]
        The unlearning process effectively retains the unimodal knowledge if it minimizes the discrepancy between the unimodal representations produced by the unlearned model $f'$ and the original model $f$:
    \begin{equation}
        \mathop{\mathbb{E}}_{(I_i, T_i) \in D_f}
        \Big[
        \psi\big(f'_I(I_i), f'_T(T_i)\big) - 
        \psi\big(f_I(I_i), f_T(T_i)\big)
        \Big] = \epsilon,
        \label{eq:propthree}
    \end{equation}
\end{definition} 
\noindent where $f_I(\cdot)$ and $f_T(\cdot)$ generate unimodal representations for image and text data respectively, the readout function $\psi$ is a vector combination operator (such as concatenation), and $\epsilon$ is an infinitesimal constant. Not that, although \cref{eq:propthree} can be applied to all training data $D_\mathrm{train}$, we only apply it to $D_f$ samples for efficiency purpose. In fact, we expect unimodal knowledge on remaining data $D_r$ to be preserved in $f'$ due to \proptwo, discussed above.

Thus, we minimize the following gap to realize \propthree: 
\begin{equation}
    \mathcal{L}_{\text{UKR}} = \text{Dis}
    \Bigg(
    \Big \{ \big[f'_I(I_i); f'_T(T_i)\big] | (I_i, T_i) \in D_f \Big\},  
    \Big \{ \big[f_I(I_i); f_T(T_i)\big] | (I_i, T_i) \in D_f\Big\}
    \Bigg),
\end{equation} where $[;]$ denotes vector concatenation. This loss aims to retain the core unimodal knowledge during training even after unlearning certain relationships. 


We note that an alternative approach is to use the fusion module $f_F$, while freezing the unimodal encoders $f_I$ and $f_T$. However, this can be limiting, especially for models like CLIP~\cite{radford2021learning}, which use nonparametric fusion modules (e.g., dot product) for modality interaction. There is also a risk that an adversarial agent might exploit the original $f_F$  and can take advantage of the frozen image and text representations. Therefore, we advocate for the strategy in $\mathcal{L}_{\text{UKR}}$ but encourage the adjustments to be minimal.

\subsection{Optimization} 
The above loss functions correspond to different key properties for multimodal unlearning. We integrate them through the following aggregate loss function, which is optimized through stochastic gradient descent:
\begin{equation}\label{eq:loss}
\mathcal{L} = 
\alpha \mathcal{L}_{\text{MD}} + 
\beta \mathcal{L}_{\text{MKR}} +
\gamma \mathcal{L}_{\text{UKR}}, 
\end{equation}
where $\mathcal{L}_{\text{MD}}$ realizes \propone by ensuring that relations between data pairs marked for deletion are unrelated by the model,
$\mathcal{L}_{\text{MKR}}$ realizes \proptwo by preserving the model's overall multimodal knowledge, and 
$\mathcal{L}_{\text{UKR}}$ realizes \propthree by preserving the model's knowledge of individual modalities. 
This aggregated loss function effectively unlearns specific data points while maintaining the general functionality and knowledge of the original model. This balanced approach is crucial for the practical application of machine unlearning, particularly in settings where both data unlearning and model performance are of importance.

\section{Experimental Setup}

\subsubsection{Tasks and Datasets}
\method is flexible and broadly applicable to a wide range of tasks, including generative ones. In this paper, we focus our evaluation on several image-text and graph-text tasks across several datasets.

\begin{itemize}
    \item \textbf{Image-Text Retrieval (TR)~and~(IR)} are the tasks of retrieving the top-k relevant texts for a given image query (TR), and, vice versa, retrieving the top-k relevant images for a given text query (IR). We use Flickr30K dataset~\cite{bojchevski2018deep} for IR and TR.
    \item \textbf{Visual Entailment (VE)} is an image-text entailment task, where the objective is to determine whether a given text hypothesis $T_i$ entails, contradicts, or is neutral with respect to a given image premise $I_i$. We use SNLI-VE dataset~\cite{xie2019visual} for VE.
    \item \textbf{Natural Language for Visual Reasoning (NLVR)} is the binary classification task of predicting whether a given text $T_i$ accurately describes a given pair of images $(I_{i,1}, I_{i,2})$. We use NLVR\textsuperscript{2} dataset~\cite{suhr-etal-2019-corpus} for NLVR. 
    \item \textbf{Graph-Text Classification} is the task of classifying whether a text indicates a specific (e.g. causal) relationship between two given entities in a subgraph. 
    We use PGR dataset~\cite{sousa-etal-2019-silver}, in which the target entities are phenotypes and genes, and the task is to determine if their relationship, as described by the accompanying text, is causal or non-causal. 
    
\end{itemize}

\begin{table}[t]
\scriptsize
\centering
\setlength{\tabcolsep}{3.9pt}
\renewcommand{\arraystretch}{0.9}
\caption{Statistics of evaluated datasets.}
\begin{tabular}{l|c|c|c|c}
\toprule
    Dataset &  Flickr30k & SNLI-VE & NLVR\textsuperscript{2} & PGR \\
    \midrule
    \# images or graphs ($I$) & 29.0K & 29.8K & 51.6K & 4.0K \\
    \# texts  ($T$) & 144.5K & 462.7K &  22.8K & 4.0K \\
    \# $I-T$ pairs & 145.0K & 529.5K & 86.4K & 4.0K \\

\bottomrule
\end{tabular}
\label{tab:data_stat}
\end{table}
 
\subsubsection{Baselines}
We compare \method to the following models:
\begin{itemize}
    \item \textbf{\retrain} is retraining a new model $f$ of same architecture from scratch with the remaining data $D_f$. 
    \item \textbf{\ft}~\cite{Golatkar2020EternalSO} is an optimization-based and modality-agnostic approach that unlearns data through continued fine-tuning. Specifically, it fine-tunes $f$ on $D_f$ with a larger learning rate, similar to catastrophic forgetting.
    \item \textbf{\neggrad}~\cite{Golatkar2020EternalSO} is an optimization-based and modality-agnostic approach that unlearns data using negative gradient. Specifically, it optimizes the original loss function of training $f$ on $D_f$ but reverses the direction of gradients to unlearn these samples. 
    \item \textbf{\dtd}~\cite{neel2021descent}, Descent to Delete is a weight scrubbing-based and modality-agnostic approach to unlearning. It assumes that the weights of $f'$ are close to the weights of $f$, trains $f$ for a few more steps while adding Gaussian noise to scrub the weights.
    \item \textbf{\codec}~\cite{Mehta2022DeepUV} is a weight scrubbing-based and uni-modal (vision only or text only) approach that approximates the Hessian matrix and performs a Newton update step to scrub the parameters while adding noise to them.
    \item \textbf{\erm}~\cite{Lin2023ERMKTPKM} is a retraining-based and uni-modal (vision only) approach that unlearns data by retraining the model with extra parameters inserted after visual feature maps to entangle correlations between classes. This method has been developed for machine unlearning in image classification.
    \item \textbf{\knowul}~\cite{jang-etal-2023-knowledge} is an optimization-based and unimodel (text only) approach that unlearns data by maximizing the log likelihood of samples in $D_f$. This method has been developed for machine unlearning in language models.
\end{itemize}

\noindent For each method, we also consider a variant where only the parameters of the fusion module $f_F$ are updated during unlearning, while the rest of the model remains frozen; this is inspired by recent works on Parameter-Efficient Tuning~\cite{chen-yang-2023-unlearn,cheng2023gnndelete,su-etal-2023-exploring}. We denote this setting by adding `\textbf{-F}' to method names. 

\subsubsection{Evaluation}
Following previous works~\cite{wang-etal-2023-kga,cheng2023gnndelete,fan2024salun,jia2023model}, we employ several standard metrics to evaluate the unlearning efficacy of different methods:
\begin{itemize}
    \item \textbf{Test Set Performance} (\dtest $\uparrow$), which evaluates the unlearned model on the original test set \dtest. We follow previous work~\cite{albef,pmlr-v162-li22n,vakil-amiri-2022-generic} to use mean recall (recall@1, recall@3, recall@10) for retrieval tasks and accuracy for other tasks. Higher values indicate that the model maintains better performance on the test set post unlearning.
    \item \textbf{Deletion Set Performance} (\df $\downarrow$), which evaluates the unlearned model on the forget set \df. The metrics are the same as those on \dtest, with the difference that lower values indicate better performance. 
    \item \textbf{Membership Inference Vulnerability} (MI $\uparrow$), which measures vulnerability against membership inference (MI) attacks. Following previous work~\cite{Golatkar2020EternalSO,cheng2023gnndelete}, we evaluate the unlearned model $f'$ in a blackbox MI setting, where the adversarial agent only has access to the output distribution of $f'$. An SVM classifier is trained using validation data as negative samples and a similarly sized subset of training data as positive samples~\cite{Golatkar2020EternalSO}. We probe the existence probability of the deleted data \df with the MI attacker before and after unlearning, and report the ratio of prior-to-post existence probabilities. A lower MI ratio indicates higher robustness to MI attacks and better protection to the data marked for deletion.
\end{itemize}

\subsubsection{Settings}
We assume that only the associations between $(I_i, T_i)$ are deleted, while individual $I_i$ and $T_i$ are not removed. For each dataset, we first train a corresponding model with the full training set (\dtrain). Then we randomly sample 5K data points from \dtrain to create our deletion set \df. We evaluate the unlearning methods across a range of deletion volumes, from 1K--5K samples with step size of 1K; this \df/\dtrain ratio matches the ratios used in previous studies~\cite{cheng2023gnndelete,chien2022certified}. \cref{tab:data_stat} shows statistics of the datasets and deleted data.
We set $\alpha, \beta, \gamma = 1$ in \cref{eq:loss}. The original models $f$ are trained until convergence before being used for deletion experiments. For deletion, we select the best checkpoint using validation set of each dataset. Supplementary materials provides additional details.\looseness-1

\subsubsection{Multimodal Architectures}
For image-text tasks, we use two popular pretrained vision-language transformers, ALBEF~\cite{albef} and BLIP~\cite{pmlr-v162-li22n}, and follow their training and evaluation settings.
For PGR, we employ the GCN~\cite{kipf2017semisupervised} and BERT~\cite{devlin-etal-2019-bert} models used in~\cite{vakil-amiri-2022-generic} to obtain unimodal subgraph and text representations respectively. These representations are then fused using a feed-forward network to obtain multimodal representations~\cite{vakil-amiri-2022-generic}. The unimodal and multimodal representations are concatenated and fed into a classifier for prediction.

\begin{table*}[t]
\scriptsize
\centering
\caption{Experimental results on image-text and graph-text datasets on ALBEF~\cite{albef}. Performance shows average of recall@1, recall@3, recall@10 on Flickr30K-TR and Flickr30K-IR, and accuracy on other datasets. 
\erm inserts trainable parameters to the vision encoder, causing the \erm-F variant ('-F' suffix in model titles denotes variants where only fusion module parameters are updated during unlearning) inapplicable. The best results are in \textbf{bold} and the second best results are \underline{underlined}. The \colorbox{lightgray}{\retrain} performance is provided \textit{for reference purpose only}. See supplementary materials for additional results.}

\begin{tabular}{l|cccc|cc|cc|cc|cc}
\toprule
\multirow{4}{*}{\textbf{Method}} & \multicolumn{8}{c|}{\textbf{Image-Text}} & \multicolumn{2}{c|}{\textbf{Graph-Text}} & \multicolumn{2}{c}{\multirow{4}{*}{\textbf{Avg.}}} \\
\cmidrule{2-11}
& \multicolumn{4}{c|}{\textbf{Flickr30K}} & \multicolumn{2}{c|}{\multirow{2}{*}{\textbf{SNLI-VE}}} & \multicolumn{2}{c|}{\multirow{2}{*}{\textbf{NLVR}\textsuperscript{2}}} & \multicolumn{2}{c|}{\multirow{2}{*}{\textbf{PGR}}} \\
& \multicolumn{2}{c}{\textbf{IR}} & \multicolumn{2}{c|}{\textbf{TR}} & & & & & & & & \\
    \cmidrule{2-13}
                & \dtest & \df & \dtest & \df & \dtest & \df & \dtest & \df & \dtest & \df & \dtest & \df \\
    \toprule
    \rowcolor{Gray}
    \retrain   & 97.8 & 50.4 & 93.5 & 50.4 & 79.4 & 50.2 & 80.3 & 50.3 & 67.5 & 50.2 & 83.4 & 50.3 \\
    \midrule
    \ft        & 96.7 & 50.4 & 94.1 & 50.4 & 79.1 & 50.5 & 80.3 & 49.8 & 67.4 & 49.9 & 83.5 & 50.2 \\
    \ft-F      & \textbf{97.1} & 49.9 & \textbf{94.6} & 49.9 & 79.5 & 49.9 & 81.2 & 50.0 & 67.5 & 50.1 & \underline{83.8} & 49.9 \\
    \neggrad   & 92.4 & 50.5 & 91.7 & 50.5 & 77.8 & 48.6 & 77.3 & 50.6 & 63.4 & 49.6 & 80.5 & 50.0 \\
    \neggrad-F & 93.3 & 50.2 & 90.6 & 50.2 & \underline{79.6} & 50.6 & \textbf{80.8} & 50.0 & 63.5 & 49.9 & 81.5 & 50.2 \\
    \dtd       & 10.3 & 51.4 &  8.9 & 51.4 & 45.2 & 50.1 & 50.8 & 49.8 & 50.0 & 50.2 & 33.0 & 50.5 \\ 
    \dtd-F     & 22.5 & 50.9 & 20.7 & 50.9 & 48.6 & 49.8 & 50.9 & 49.8 & 53.6 & 50.2 & 39.2 & 50.2 \\
    \codec     & 83.5 & 50.0 & 78.5 & 50.0 & 56.7 & 49.9 & 55.3 & 52.7 & 57.8 & 48.8 & 66.3 & 50.3 \\
    \codec-F   & 87.4 & 49.4 & 50.6 & 48.2 & 57.4 & 48.4 & 56.8 & 53.1 & 59.1 & 46.9 & 62.2 & 49.2 \\
    \erm       & 57.4 & 48.7 & 56.2 & 49.0 & 53.2 & 48.9 & 52.9 & 50.8 & \multicolumn{2}{c|}{N/A} & 54.9 & 49.3 \\
    \erm-F     & \multicolumn{12}{c}{N/A} \\
    \knowul    & 95.1 & 50.4 & 90.3 & 50.4 & 75.7 & 49.8 & 76.3 & 50.4 & 64.8 & 49.7 & 80.4 & 50.2  \\
    \knowul-F  & 94.4 & 50.2 & 94.1 & 50.2 & 79.1 & 49.7 & 76.8 & 50.4 & 66.1 & 48.8 & 82.1 & 49.8 \\
    \midrule
    \method    & \textbf{97.1} & \textbf{33.2} & \underline{94.3} & \textbf{33.2} & \textbf{79.8} & \textbf{35.3} & \textbf{80.8} & \textbf{23.5} & \textbf{68.5} & \textbf{18.6} & \textbf{84.2} & \textbf{28.7} \\
    \method-F  & \underline{96.8} & \underline{34.4} & 94.1 & \underline{34.5} & 79.5 & \underline{36.3} & \underline{80.4} & \underline{26.4} & \textbf{67.7} & \textbf{19.5} & 83.7 & \underline{30.2} \\
\bottomrule
  \end{tabular}
  \label{tab:main}
\end{table*}

\section{Main Results}
The results in \cref{tab:main} show that on average across all {\em image-text} tasks, \method achieves $88.0$ on \dtest, outperforming all baselines by $17.4$ absolute points. Furthermore, it achieves $31.1$ on \df, outperforming all baselines by $17.6$ absolute points. In addition, \method effectively reduces the likelihood of deleted data (\df) being identified, resulting in an average MI ratio of $1.3$ across all tasks. These results indicate that \method can accomplish effective and targeted unlearning, while maintaining strong capability and utility on downstream tasks. 
In addition, on PGR dataset, \method outperforms baselines on \dtest, outperforming \ft by $+1.1$ points, \neggrad by $+5.1$ points, \dtd by $+38.5$ points, \codec by $+10.7$ points, and \knowul by $+3.7$ points. In addition, \method achieves $18.6$ on deleted data (\df), significantly outperforming the top-performing baseline model, \codec ($48.8$). The performance of \method on the original test set \dtest is better that that of \retrain by $+1.0$, see \cref{tab:main}.\looseness-1

\subsubsection{Comparison to Modality-agnostic Approaches}
Results in \cref{tab:main} show that none of the existing modality-agnostic approaches is sufficient to unlearn multimodal samples from trained models. Specifically, on \dtest, \method outperforms \ft, \neggrad, \dtd, \codec by $+0.6$, $+3.3$, $+49.3$ and $+19.6$ average points respectively. The corresponding improvements on \df are $+19.0$, $+18.8$, $+19.4$ and $+19.4$ points respectively. The lower performance of these approaches show that they can't remove learned multimodal dependencies.\looseness-1

\subsubsection{Comparison to Unimodal Approaches}
Results in \cref{tab:main} show that unimodal unlearning approaches do not effectively translate to multimodal contexts. Specifically, \method outperforms \erm and \knowul by substantial margins of $+33.2$ and $+3.8$ absolute points on \dtest, $+18.0$ and $19.0$ absolute points on \df. Updating the knowledge on one of the modalities results in drop on both test set performance and model's ability in forgetting \df. Therefore, merely unlearning a single modality is inadequate for comprehensive unlearning in multimodal settings, where removal of inter-modality association is anticipated.


\subsubsection{Limitations of Scrubbing Methods and \retrain} 
Our results in \cref{tab:main} show that scrubbing methods, despite with theoretical guarantee, fall short in multimodal unlearning in practice; the scrubbing methods \dtd and \codec achieve an average performance of $37.6$ and $59.5$, respectively, which are considerably lower than that of \ft ($68.5$) and \neggrad ($67.5$). They also result in extremely low performance of $26.15$ and $68.5$ on \dtest respectively. In case of multimodal settings, we argue that scrubbing or noise addition disrupts the original learned dependencies, particularly when model parameters are shared, e.g. by nodes in graphs~\cite{cheng2023gnndelete} or by different fused modalities. In unimodal settings, where scrubbing methods are tested, since the encoder encompasses most of the model parameters, scrubbing methods do not show strong influence on the performance of downstream tasks. 
\method even outperforms \retrain by $xx$ and $xx$ absolute point on \dtest and \df, respectively. As several previous works have observed~\cite{fan2024salun,unbound,goel2022evaluating,cheng2023gnndelete}, \retrain does not necessarily serve as the gold standard for unlearning.
These results indicate that matching model parameters does not necessarily mean successful unlearning due to potential distribution discrepancy in model parameters, as noted by many previous works~\cite{thudi2021necessity,fan2024salun,unbound,goel2022evaluating,cheng2023gnndelete}.


\subsubsection{Membership Inference Attack}
\method achieves a reduced probability of detecting deleted data (\df) compared to before unlearning (see the prior-to-post MI ratio in \cref{tab:main}). This indicates that \method can better protect the deleted data and is less susceptible to MI attacks. Specifically, \method outperforms non-scrubbing baselines (\ft, \neggrad, \erm, \knowul) by $0.19$ absolute points in MI ratio. We note that, although scrubbing methods like \dtd and \codec show a significant decrease in existence probability (higher MI ratios) than non-scrubbing methods, the drop applies to all data including both \dr and \df. This shows that the unlearning of scrubbing methods is not targeted at a specific subset of data, but the entire data, which signals a failed deletion.

\subsubsection{All Key Properties Contribute to Unlearning}
Through an ablation study, we assess the individual contributions of the key properties proposed in \method: \propone (MD), \proptwo (UKR), and \propthree (MKR). \cref{tab:abl} shows that excluding MD 
results in a significant decline in model's ability in distinguishing between \df and \dr, with performance dropping from $76.5$ to $50.3$ ($-26.2$). 
Both UKR and MKR serve as objectives for maintaining the original knowledge acquired by the model, targeting at \df and \dr respectively. 
The exclusion of UKR and MKR lead to performance drops of $0.5$ and $6.6$ on downstream tasks respectively. The more substantial impact observed by removing MKR can be attributed to two factors: 
(1) \dr usually has a much larger size than \df, leading to a much larger influence for MKR; and 
(2) downstream tasks tend to rely more heavily on multimodal knowledge than unimodal knowledge, making MKR crucial for maintaining model performance.

\subsubsection{Utility of Unimodal Knowledge}
We highlight that the multimodal unlearning in \method can retain the utility of unimodal embeddings. We train unimodal image and text classifiers $g_I$ on $f(I)$, and test the classifiers with the unimodal embeddings after unlearning, namely $f'(I)$. This experiment can show how much unlearning preserves the utility of unimodal embeddings $f'(I)$ after unlearning, where an optimal unlearning method should yield same performance for $g_I(f(I))$ and $g_I(f'(I))$. \cref{tab:ukr} \method outperforms the best baseline (\ft) by $+0.4$ in accuracy, while removing UKR in optimization leading to dramatic drop by $-5.7$.


\begin{table*}[t]
\scriptsize
\centering
\caption{Prior-to-post MI ratio on image-text datasets with the best-performing baseline. The \colorbox{lightgray}{\retrain} performance is provided \textit{for reference purpose only}. See supplementary materials for additional results.}
\begin{tabular}{l|c|c|c|c|c|c}
\toprule
\multirow{2}{*}{\textbf{Method}} & \multicolumn{4}{c|}{\textbf{Image-Text}} & \textbf{Graph-Text} & \multirow{2}{*}{\textbf{Avg.}} \\
\cmidrule{2-6}
 & \textbf{Flickr-IR} & \textbf{Flickr-TR} & \textbf{SNLI-VE} & \textbf{NLVR}\textsuperscript{2} & \textbf{PGR} & \\
    \toprule
    \rowcolor{Gray}
    \retrain   & 1.10 & 1.10 & 1.05 & 1.07 & 1.09 & 1.08 \\
    \midrule
    \codec     & 1.21 & 1.21 & 1.23 & 1.23 & 1.07 & 1.19 \\ 
    \codec-F   & 1.22 & 1.22 & 1.26 & 1.26 & 1.09 & 1.21 \\ 
    \midrule
    \method    & \textbf{1.27} & \textbf{1.27} & \textbf{1.30} & \textbf{1.25} & \textbf{1.24} & \textbf{1.27} \\
    \method-F  & \underline{1.25} & \underline{1.25} & \underline{1.26} & \underline{1.21} & \underline{1.20} & \underline{12.4} \\
\bottomrule    
  \end{tabular}
  \label{tab:mi}
\end{table*}

\begin{table}[t]
\scriptsize
\centering
\caption{Ablation study of unlearning objectives of \method. All three objectives contributes to both downstream performance on \dtest and forgetting \df.}
\begin{tabular}{l|cc|cc}
\toprule
               & \multicolumn{2}{c|}{\textbf{NLVR\textsuperscript{2}}} & \multicolumn{2}{c}{\textbf{PGR}} \\
                \cmidrule{2-5}
               & \dtest & \df & \dtest & \df \\
    \midrule
    \rowcolor{Gray}
    \retrain   & 80.3 & 50.3 & 67.5 & 50.2 \\
    \midrule
    Full model & \textbf{80.8} & \textbf{23.5} & \textbf{67.8} & 1\textbf{8.6} \\
    \midrule
    - MD       & 80.3 & 50.3 & 67.5 & 49.3 \\ 
    - UKR      & 79.2 & 25.8 & 66.3 & 22.6 \\ 
    - MKR      & 77.1 & 25.6 & 64.8 & 23.7 \\ 
\bottomrule
\end{tabular}
\vspace{-10pt}
\label{tab:abl}
\end{table}

\begin{table*}[t]
\scriptsize
\centering
\caption{Accuracy of image classification using unimodal embeddings after unlearning.}
\begin{tabular}{l|c|c|c|c|c|c|c}
\toprule
 & \textbf{\ft} & \textbf{\neggrad} & \textbf{\dtd} & \textbf{\codec} & \textbf{\knowul} & \textbf{\method} & w/o UKR \\
\midrule
\textbf{Acc.}   & 83.2 & 81.7 & 43.8 & 55.2 & 82.7 & \textbf{83.6} & 77.9 \\
\bottomrule    
  \end{tabular}
  \label{tab:ukr}
\end{table*}

\subsubsection{Updating All Parameters vs. Fusion Module Only}
We experiment whether updating all model parameters or just those of the modality fusion module is more effective for unlearning. We denote these two approaches as \textsc{Method} (updating all parameters) and \textsc{Method}-F (updating only the fusion module parameters). \textbf{For \method}, focusing solely on updating $f_F$ (the fusion module) is somewhat akin to bypassing the optimization for $\mathcal{L}_{\mathrm{UKR}}$, though not exactly the same. We find that this version of \method exhibits less fluctuation in  performance on \dtest during training but tends to converge more slowly on \dfauc compared to the full version of \method. 
\textbf{For scrubbing-based methods (\dtd, \codec)}, updating all the parameters results in a complete loss previously acquired knowledge, resulting in random performance across all tasks. Conversely, targeting only the fusion modules for scrubbing helps retain performance on downstream tasks. This suggests that 
(a) robust unimodal knowledge plays a critical role in multimodal tasks, and 
(b) the fusion module is more resilient to noise or minor perturbations than the unimodal encoders. However, neither approach significantly aids the model in distinguishing \df from \dr or in protecting \df against MI attacks.
\textbf{For modality-agnostic approaches}, we observe negligible differences between the two strategies, with a marginal performance gap of less than $0.6$ absolute points. This indicates that for these methods, the strategy chosen for parameter updating has minimal impact on overall performance.

\begin{figure}[t]
    \centering
    \includegraphics[width=0.9\textwidth]{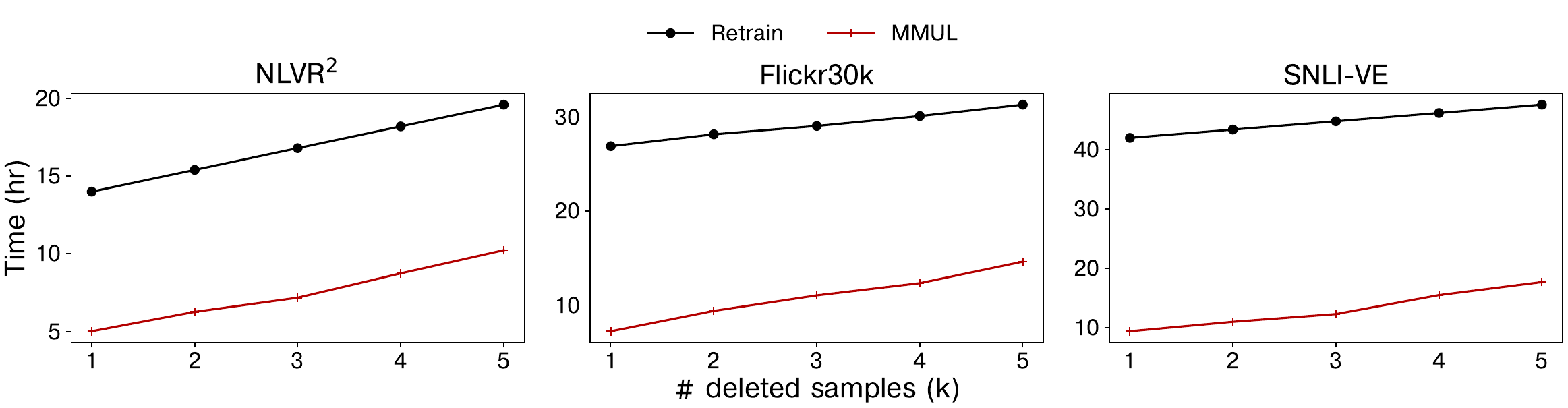}
    \caption{Training time of unlearning methods.}
    \label{fig:time}
    \vspace{-10pt}
\end{figure}

\subsubsection{Efficiency}
{\em For training time,} \cref{fig:time} presents a comparative analysis of the training times for \method and \retrain across datasets with an increasing size of $|$\df$|$. The results indicate that \method is efficient to run, and exhibits a linear growth in running time as $|$\df$|$ increases. This trend illustrates \method's scalability and effectiveness in managing larger volumes of data marked for deletion. 
{\em For trainable parameters,} several works have optimized unlearning in an parameter-efficient manner~\cite{cheng2023gnndelete,fan2024salun,chen-yang-2023-unlearn}, similar to parameter-efficient fine-tuning~\cite{peft,su-etal-2023-exploring}. We argue that \method-F only optimizes a small portion of the parameters, while delivering comparable performance as the full model with trivial gap.

\subsubsection{Why \method works}
We attribute the superior performance of \method to the three proposed properties. MD aims at relaxing the dependencies between $(I_i, T_i)$ pairs, while MKR and UKR preserve the acquired knowledge. Compared to existing unimodal approaches, MD can remove the relationships between data modalities. Compared to existing modality-agnostic approaches, MKR and UKR maintains the capability of model on multimodal tasks. Collectively, all three properties contribute to the success of unlearning, where strong downstream performance and successful targeted forgetting are desired in the same time. 

\section{Related work}
\subsubsection{Machine Unlearning} 
Existing machine unlearning research can be categorized into four classes. (a): \textit{Efficient retraining methods} retrain models on partitioned data and learn to combine the predictions from each sharded model to maintain performance~\cite{bourtoule2021machine,Wu2020DeltaGradRR,pmlr-v119-wu20b,9796721,Chen2021GraphU,Dukler_2023_ICCV}. Training on partitioned multimodal data may inversely affect the dependencies between modalities, lead to overfitting, be inefficient and not scalable.
\textit{(b) Weight scrubbing methods} adopt a one-shot weight update followed by added noise to model weights, whose probability distribution is indistinguishable from that of a model retrained from scratch with theoretical guarantee~\cite{Golatkar2020EternalSO,guo-2020-certified-removal,neel2021descent,pmlr-v139-brophy21a,ceu,izzo2021approximate,suriyakumar2022algorithms,liu2023certified}. However, the assumption of strongly convex loss cannot be guaranteed in multimodal setting. Added noise may affect the modality dependencies of all data points and poor empirical performance~\cite{thudi2021necessity}.
\textit{(c) Teacher-student unlearning:} centers on making the student (unlearned model) to follow the teacher on deleted data. The teacher can be a separate model~\cite{Wang2023KGAAG}, the opposite direction of the original model~\cite{unbound}, an untrained model~\cite{Chundawat2022CanBT}, random node pairs~\cite{cheng2023gnndelete}, a random label~\cite{fan2024salun}, an error matrix~\cite{fast-yet}.

\subsubsection{Unlearning in Vision}
Certified data removal~\cite{guo-2020-certified-removal} views image classifiers as frozen feature extractors and linear classifiers and derives a Newton update step to remove information. Boundary Unlearning~\cite{Chen2023BoundaryUR} tackles the problem of unlearning an entire class by shifting the decision boundary, either shrinking or expanding the original boundary. ERM-KTP~\cite{Lin2023ERMKTPKM} inserts entanglement-reduced mask (ERM) layers and retrains the model,  incorporated during training and doesn't handle existing trained models. MUter~\cite{Liu_2023_ICCV} proposes a close-form solution for unlearning on adversarially trained vision models based on influence measure.

\subsubsection{Unlearning in Language}
KGA~\cite{wang-etal-2023-kga} aligns the knowledge gap between the deleted data and remaining data of the unlearned model with a separate model trained with extra data. However, obtaining high quality data from the same distribution may not be easy, especially in multimodal setting. Other methods include negative gradient~\cite{jang-etal-2023-knowledge}, encouraging model to generate dissimilar text to ground truth~\cite{kassem-etal-2023-preserving}, making samples unlearnable~\cite{li-liu-2023-make}. EUL~\cite{chen-yang-2023-unlearn} maximizes the difference between the unlearning model and original model on the deleted samples.

\subsubsection{Unlearning in Graphs}
GraphEraser~\cite{chen2021graph} re-trains multiple models on partitioned graphs, which removes structural information and hurt performance. CGU~\cite{Chien2023EfficientMU}, CEU~\cite{ceu} and \textsc{Projector}~\cite{pmlr-v206-cong23a} assume linear underlying models, which limits their practical performance and applicability. GNNDelete~\cite{cheng2023gnndelete} performs local update to fulfill Deleted Edge Consistency and Neighborhood Influence. D2DGN~\cite{sinha2023distill} formulates unlearning as distillation from the original model.

\subsubsection{Other Unlearning Work}
Other approaches include sparsity~\cite{jia2023model,learn-to-forget,fan2024salun} and operations on gradient~\cite{pmlr-v134-ullah21a,Hoang_2024_WACV}, training with ability to unlearn in mind~\cite{zhang2022prompt,Lin2023ERMKTPKM}, in continual learning setting~\cite{Liu2022ContinualLA}, for recommender systems~\cite{10.1145/3485447.3511997,li2022making,li2023ultrare}, in Bayesian models~\cite{nguyen2020variational}, without accessing training data~\cite{chundawat2022zero}, for shallow models~\cite{10.1145/3448016.3457239,ginart2019making}, handling a sequence of deletion requests~\cite{NEURIPS2021_87f7ee4f}, for regression models~\cite{pmlr-v202-tarun23a}, verification~\cite{sommer2020towards}, vulnerability to attack~\cite{zhao2023static}, under federated setting~\cite{pan2023machine}, trade-off with reverting decisions~\cite{pawelczyk2023on}, and benchmarking~\cite{cheng2024mu,maini2024tofu}. Applications of unlearning include removing bias~\cite{setlur2022adversarial,chen2023fast}, text generation~\cite{lu2022quark}, alleviating backdoor attack~\cite{wei2023shared}, and conducting data poison attack~\cite{di2023hidden}..



\section{Conclusion}
This work formulates the problem of multimodal unlearning 
and introduces \method--the first multimodal unlearning approach that is task and architecture agnostic, and is efficient to use. 
It implements three key properties for effective multimodal unlearning: 
(a): \propone, which aims to decouple the association between individual unimodal data points marked for deletion, rendering them as unrelated data points,
(b): \proptwo, which retains the multimodal representation capability of the model post-unlearning, and
(c): \propthree, which retains the unimodal representation capability of the model post-unlearning.
Through extensive experiments across image-text and graph-text tasks and several datasets and architectures, we show that \method reliably unlearn multimodal relationships, and outperforms existing modality-agnostic and unimodal unlearning methods in maintaining downstream performance, and distinguishing between deleted data and remaining data. In addition, we show that the model can maintain the multimodal and unimodal knowledge post unlearning, 
can provide better protection to unlearned data, 
and is robust against adversarial attacks.

%
%
\bibliographystyle{splncs04}
\bibliography{reference}

\newpage
\appendix
\begin{table*}
\tiny
\centering
\setlength{\tabcolsep}{1.7pt}
\caption{Experimental results on image-text and graph-text datasets on BLIP~\cite{pmlr-v162-li22n}. Performance shows average of recall@1, recall@3, recall@10 on Flickr30K-TR and Flickr30K-IR, and accuracy on other datasets. \erm inserts trainable parameters to the vision encoder, causing the \erm-F variant ('-F' suffix in model titles denotes variants where only fusion module parameters are updated during unlearning) inapplicable. The best results are in \textbf{bold} and the second best results are \underline{underlined}. The \colorbox{lightgray}{\retrain} performance is provided \textit{for reference purpose only}.}
\begin{tabular}{l|cccc|cc|cc|cc}
\toprule
\multirow{3}{*}{\textbf{Method}} & \multicolumn{4}{c|}{\textbf{Flickr30K}} & \multicolumn{2}{c|}{\multirow{2}{*}{\textbf{SNLI-VE}}} & \multicolumn{2}{c|}{\multirow{2}{*}{\textbf{NLVR}\textsuperscript{2}}} & \multicolumn{2}{c}{\multirow{2}{*}{\textbf{Avg.}}} \\
& \multicolumn{2}{c|}{\textbf{IR}} & \multicolumn{2}{c|}{\textbf{TR}} & & & & & & \\
    \cmidrule{2-11}
                & \dtest & \df & \dtest & \df & \dtest & \df & \dtest & \df & \dtest & \df \\
    \toprule
    \rowcolor{Gray}
    \retrain   & 98.6 & 50.5 & 94.1 & 50.5 & 79.7 & 50.2 & 79.1 & 50.4 & 87.8 & 50.4 \\
    \midrule
    \ft        & 98.7 & 50.4 & 94.1 & 50.4 & 80.1 & 50.3 & 80.1 & 50.2 & 88.2 & 50.3 \\
    \ft-F      & 97.8 & 50.1 & 94.6 & 50.1 & 80.3 & 50.2 & 80.6 & 50.1 & 88.3 & 50.1 \\
    \neggrad   & 94.7 & 50.0 & 92.4 & 50.0 & 78.5 & 49.9 & 78.0 & 50.3 & 85.9 & 50.0 \\
    \neggrad-F & 94.1 & 49.8 & 92.0 & 49.8 & 78.8 & 49.6 & 79.6 & 50.1 & 86.1 & 49.8 \\
    \dtd       & 11.5 & 50.0 & 10.8 & 50.0 & 49.3 & 50.0 & 50.2 & 50.0 & 30.4 & 50.0 \\ 
    \dtd-F     & 18.3 & 47.4 & 17.9 & 47.6 & 57.4 & 46.6 & 51.1 & 48.6 & 36.1 & 47.5 \\
    \codec     & 86.1 & 49.4 & 81.0 & 49.6 & 60.3 & 48.8 & 58.2 & 45.3 & 71.3 & 48.2 \\
    \codec-F   & 87.3 & 48.2 & 82.3 & 48.2 & 62.7 & 47.2 & 59.3 & 42.5 & 72.9 & 46.5 \\
    \erm       & 57.7 & 50.1 & 56.4 & 50.1 & 52.9 & 50.5 & 53.4 & 46.1 & 55.1 & 49.1 \\
    \erm-F     & \multicolumn{10}{c}{N/A} \\
    \knowul    & 94.8 & 50.5 & 93.3 & 50.5 & 74.5 & 50.4 & 75.4 & 49.0 & 84.5 & 50.1 \\
    \knowul-F  & 94.2 & 50.4 & 93.1 & 50.4 & 73.8 & 50.6 & 75.6 & 49.3 & 84.1 & 50.1 \\
    \midrule
    \method    & 98.7 & 32.7 & 95.3 & 32.7 & 80.7 & 40.7 & 80.8 & 25.1 & 88.9 & 32.8 \\
    \method-F  & 98.0 & 33.9 & 94.7 & 33.9 & 80.5 & 40.5 & 80.7 & 26.5 & 88.5 & 33.7 \\
\bottomrule    
  \end{tabular}
  \label{tab:blip}
\end{table*}

\begin{table}
\tiny
\centering
\caption{Prior-to-post ratio of existence probability of \df given by a MI attacker on ALBEF. The best performance is in \textbf{bold} and the second best is \underline{underlined}. In addition, the \colorbox{lightgray}{\retrain} performance is provided \textit{for reference purpose only}. Results on BLIP is shown in Supplementary material.}
\label{tab:mi_albef}
\begin{tabular}{l|c|c|c|c|c}
\toprule
\textbf{Method} & \textbf{Flickr-TR} & \textbf{Flickr-IR} & \textbf{SNLI-VE} & \textbf{NLVR}\textsuperscript{2} & \textbf{Avg.}\\
    \toprule
    \rowcolor{Gray}
    \retrain   & 1.10 & 1.10 & 1.05 & 1.07 & 1.08 \\
    \ft        & 1.03 & 1.03 & 1.04 & 1.08 & 1.04 \\
    \ft-F      & 1.06 & 1.06 & 1.07 & 1.09 & 1.07 \\
    \neggrad   & 1.11 & 1.11 & 1.09 & 1.06 & 1.09 \\
    \neggrad-F & 1.14 & 1.14 & 1.10 & 1.08 & 1.11 \\
    \dtd       & 1.41 & 1.41 & 1.60 & 1.71 & 1.53 \\ 
    \dtd-F     & 1.40 & 1.40 & 1.58 & 1.66 & 1.51 \\
    \codec     & 1.21 & 1.21 & 1.23 & 1.23 & 1.22 \\
    \codec-F   & 1.22 & 1.22 & 1.26 & 1.26 & 1.24 \\
    \erm       & 1.10 & 1.10 & 1.11 & 1.21 & 1.13 \\
    \erm-F     & \multicolumn{5}{c}{N/A} \\
    \knowul    & 0.97 & 0.97 & 1.04 & 1.07 & 1.01 \\
    \knowul-F  & 0.98 & 0.98 & 1.10 & 1.04 & 1.02 \\
    \midrule
    \method    & 1.27 & 1.27 & 1.30 & 1.25 & 1.27 \\
    \method-F  & 1.25 & 1.25 & 1.26 & 1.21 & 1.24 \\
\bottomrule    
  \end{tabular}
\end{table}

\begin{table*}[t]
\tiny
\centering
\label{tab:mi_blip}
\caption{Prior-to-post MI ratio of existence probability of \df given by a MI attacker on BLIP. The \colorbox{lightgray}{\retrain} performance is provided \textit{for reference purpose only}.}
\begin{tabular}{l|c|c|c|c|c}
\toprule
\textbf{Method} & \textbf{Flickr - TR} & \textbf{Flickr - IR} & \textbf{SNLI-VE} & \textbf{NLVR}\textsuperscript{2} & \textbf{Avg.}\\
    \toprule
    \rowcolor{Gray}
    \retrain   & 1.10 & 1.10 & 1.05 & 1.07 & 1.08 \\
    \midrule
    \ft        & 1.03 & 1.03 & 1.04 & 1.08 & 1.04 \\
    \ft-F      & 1.06 & 1.06 & 1.07 & 1.09 & 1.07 \\
    \neggrad   & 1.11 & 1.11 & 1.09 & 1.06 & 1.09 \\
    \neggrad-F & 1.14 & 1.14 & 1.10 & 1.08 & 1.11 \\
    \dtd       & 1.18 & 1.18 & 1.15 & 1.15 & 1.16 \\ 
    \dtd-F     & 1.20 & 1.20 & 1.16 & 1.16 & 1.18 \\
    \codec     & 1.21 & 1.21 & 1.13 & 1.13 & 1.17 \\
    \codec-F   & \underline{1.22} & \underline{1.22} & 1.16 & 1.16 & 1.19 \\
    \erm       & 1.10 & 1.10 & 1.14 & 1.16 & 1.12 \\
    \erm-F     & \multicolumn{5}{c}{N/A} \\
    \knowul    & 0.97 & 0.97 & 1.04 & 1.07 & 1.01 \\
    \knowul-F  & 0.98 & 0.98 & 1.10 & 1.04 & 1.02 \\
    \midrule
    \method    & \textbf{1.24} & \textbf{1.24} & \textbf{1.22} & \textbf{1.23} & \textbf{1.23} \\
    \method-F  & \underline{1.22} & \underline{1.22} & \underline{1.21} & \underline{1.21} & \underline{1.22} \\
\bottomrule    
  \end{tabular}
\end{table*}

\section{Additional results}
The results in \cref{tab:blip} show that on average across all {\em image-text} tasks on BLIP, \method achieves $88.9$ on \dtest, outperforming all baselines by $19.7$ absolute points. Furthermore, it achieves $32.8$ on \df, outperforming all baselines by $16.9$ absolute points. In addition, \method effectively reduces the likelihood of deleted data (\df) being identified, resulting in an average MI ratio of $0.13$ across all tasks. These results indicate that \method can accomplish effective and targeted unlearning, while maintaining strong capability and utility on downstream tasks. 

Similar to results on ALBEF, we observe that on BLIP none of the existing modality-agnostic approaches is sufficient to unlearn multimodal samples from trained models. Specifically, on \dtest, \method outperforms \ft, \neggrad, \dtd, \codec by $+0.7$, $3.0$, $58.5$, and $17.6$ average points respectively. The corresponding improvements on \df are $+17.5$, $+17.2$, $+17.2$ and $+15.4$ points respectively. The lower performance of these approaches show that they can't remove learned multimodal dependencies.

Results in \cref{tab:blip} show that unimodal unlearning approaches do not effectively translate to multimodal contexts. Specifically, \method outperforms \erm and \knowul by substantial margins of $+33.8$ and $+4.4$ absolute points on \dtest, $+16.3$ and $17.3$ absolute points on \df. Updating the knowledge on one of the modalities results in drop on both test set performance and model's ability in forgetting \df. Therefore, merely unlearning a single modality is inadequate for comprehensive unlearning in multimodal settings, where removal of inter-modality association is anticipated.

We also present the detailed MI results in Table~\cref{tab:mi_albef} and Table~\cref{tab:mi_blip} for ALBEF and BLIP, respectively. \method achieves a reduced probability of detecting deleted data (\df) compared to before unlearning. This indicates that \method can better protect the deleted data and is less susceptible to MI attacks. 


\section{Implementation details}
We use the LAVIS~\cite{li-etal-2023-lavis} package for implementation. We adopt a same learning rate and batch size as previous work~\cite{albef,pmlr-v162-li22n} on different backbones and downstream tasks.

\end{document}